\documentclass[a4paper,conference]{IEEEtran}
\usepackage{amsmath,epsfig}
\usepackage{cite}
\usepackage{framed,multirow,booktabs}
\usepackage{graphicx}
\usepackage{amssymb,bm,amsfonts}
\usepackage{algorithmic}
\usepackage{textcomp}
\usepackage{amsthm}
\usepackage{lineno}
\ifCLASSINFOpdf
\else
\fi

\begin{document}
\title{Trajectory-based Radical Analysis Network for Online Handwritten Chinese Character Recognition}

\author{\IEEEauthorblockN{Jianshu Zhang, Yixing Zhu, Jun Du and
Lirong Dai}
\IEEEauthorblockA{National Engineering Laboratory for Speech and Language Information Processing\\
University of Science and Technology of China,
Hefei, Anhui, P. R. China\\ Email: xysszjs@mail.ustc.edu.cn, zyxsa@mail.ustc.edu.cn, jundu@ustc.edu.cn, lrdai@ustc.edu.cn}
}

\maketitle

\begin{abstract}
  Recently, great progress has been made for online handwritten Chinese character recognition due to the emergence of deep learning techniques. However, previous research mostly treated each Chinese character as one class without explicitly considering its inherent structure, namely the radical components with complicated geometry. In this study, we propose a novel trajectory-based radical analysis network (TRAN) to firstly identify radicals and analyze two-dimensional structures among radicals simultaneously, then recognize Chinese characters by generating captions of them based on the analysis of their internal radicals. The proposed TRAN employs recurrent neural networks (RNNs) as both an encoder and a decoder. The RNN encoder makes full use of online information by directly transforming handwriting trajectory into high-level features. The RNN decoder aims at generating the caption by detecting radicals and spatial structures through an attention model. The manner of treating a Chinese character as a two-dimensional composition of radicals can reduce the size of vocabulary and enable TRAN to possess the capability of recognizing unseen Chinese character classes, only if the corresponding radicals have been seen. Evaluated on CASIA-OLHWDB database, the proposed approach significantly outperforms the state-of-the-art whole-character modeling approach with a relative character error rate (CER) reduction of 10\%. Meanwhile, for the case of recognition of 500 unseen Chinese characters, TRAN can achieve a character accuracy of about 60\% while the traditional whole-character method has no capability to handle them.
\end{abstract}


\IEEEpeerreviewmaketitle

\section{Introduction}
\label{sec:Introduction}
Machine recognition of handwritten Chinese characters has been studied for decades~\cite{suen1980automatic}. It is a challenging problem due to a large number of character classes and enormous ambiguities coming from handwriting input. Although some conventional approaches have obtained great achievements~\cite{plamondon2000online,liu2004online,zhang2017drawing,yang2016dropsample,zhong2015high}, they only treated the character sample as a whole without considering the similarity and internal structures among different characters. And they have no capability of dealing with unseen character classes.

However, Chinese characters can be decomposed into a few fundamental structure components, called radicals~\cite{chang1973interactive}. It is an intuitive way to first extract information of radicals that is embedded in Chinese characters and then use this knowledge for recognition. In the past few decades, lots of efforts have been made for radical-based Chinese character recognition. For example, \cite{wang2001optical} proposed a matching method for radical-based Chinese character recognition. It first detected radicals separately and then employed a hierarchical radical matching method to compose radicals into a character. \cite{ma2008new} tried to over-segment characters into candidate radicals while the proposed way could only handle the left-right structure and over-segmentation brings many difficulties. Recently, \cite{wang2017label} proposed a multi-label learning for radical-based Chinese character recognition. It turned a character class into a combination of several radicals and spatial structures. Generally, these approaches have difficulties when dealing with radical segmentation and the analysis of structures among radicals is not flexible. Besides, they did not focus on recognizing unseen Chinese character classes.

\begin{figure}
\centering
\includegraphics[width=3.5in]{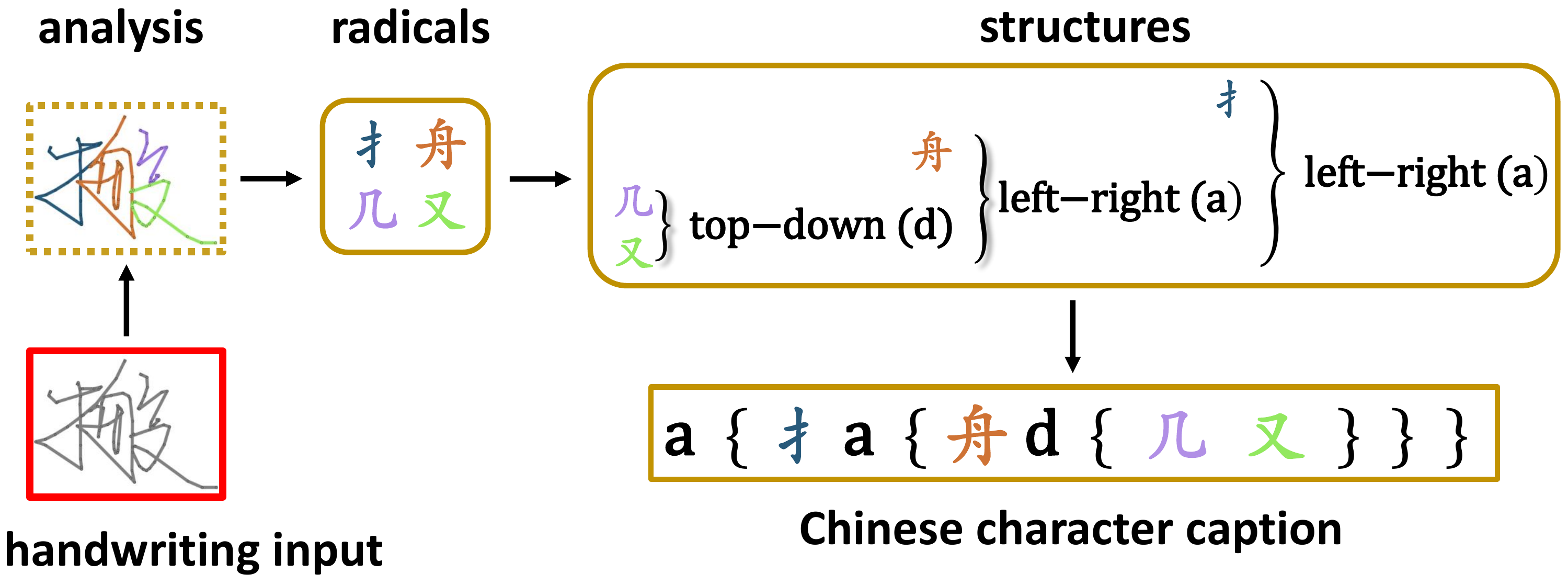}
\caption{Illustration of TRAN to recognize Chinese characters by analyzing the radicals and the corresponding structures.}
\label{fig:radical-learning-way}
\end{figure}

In this paper, we propose a novel radical-based approach to online handwritten Chinese character recognition, namely trajectory-based radical analysis network (TRAN). Different from above mentioned radical-based approaches, in TRAN the radical segmentation and structure detection are automatically addressed by an attention model which is jointly optimized with the entire network. The main idea of TRAN is to decompose a Chinese character into radicals and detect the spatial structures among radicals. We then describe the analysis of radicals as a Chinese character caption. A handwritten Chinese character is successfully recognized when its caption matches ground-truth. To be more accessible, we illustrate the TRAN learning way in Fig.~\ref{fig:radical-learning-way}. The online handwritten Chinese character input is visualized at the bottom-left of Fig.~\ref{fig:radical-learning-way}. It is composed of four different radicals. The handwriting input is finally recognized as the bottom-right Chinese character caption after the top-down and left-right structures among radicals are detected. Based on analysis of radicals, the proposed TRAN possesses the capability of recognizing unseen Chinese character classes if the radicals have been seen.

The proposed TRAN is an improved version of attention-based encoder-decoder model~\cite{bahdanau2014neural} with RNN~\cite{graves2012supervised}. The attention-based encoder-decoder model has been extensively applied to many applications including machine translation~\cite{cho2014learning,luong2015effective}, image captioning~\cite{xu2015show,vinyals2015show}, speech recognition~\cite{bahdanau2016end} and mathematical expression recognition~\cite{zhang2017watch,zhang2017multi}. The raw data of online handwritten Chinese character input are variable-length sequence (xy-coordinates). TRAN first employs a stack of bidirectional RNN~\cite{graves2013speech} to encode input sequence into high-level representations. Then a unidirectional RNN decoder converts the high-level representations into output character captions one symbol at a time. For each predicted radical, a coverage based attention model~\cite{zhang2017gru} built in the decoder scans the entire input sequence and chooses the most relevant part to describe a segmented radical or a two-dimensional structure between radicals. Our proposed TRAN is related to our previous work~\cite{zhang2017ran} with two main differences: 1) \cite{zhang2017ran} focused on the application of RAN on printed Chinese character recognition while this paper focuses on handwritten Chinese character recognition. It is interesting to investigate the performance of RAN on handwritten Chinese character recognition as handwritten characters are much more ambiguous due to the diversity of writing styles. 2) Instead of transforming online handwritten characters into static images and employing convolutional neural network~\cite{krizhevsky2012imagenet} to encode them, we choose to directly encode the raw sequential data by employing an RNN encoder in order to fully exploit the dynamic trajectory information that can not be recovered from static images.

The main contributions of this study are as follows:
\begin{itemize}
  \item We propose TRAN for online handwritten Chinese character recognition.
  \item The size of radical vocabulary is largely less than Chinese character vocabulary, leading to decrease of redundancy among output classes and improvement of recognition performance.
  \item TRAN possess the ability of recognizing unseen or newly created Chinese characters, only if the radicals have been seen.
  \item We experimentally demonstrate how RAN performs on online handwritten Chinese character recognition compared with state-of-the-arts and show its effectiveness on recognizing unseen character classes.
\end{itemize}

%The remainder of the paper is organized as follows. Section~\ref{sec:Description of Chinese character caption} describes the rule of generating captions of Chinese characters. Section~\ref{sec:The proposed approach} introduces the architecture of the proposed TRAN. In Section~\ref{sec:Experiments}, experimental results and visualization analysis are reported. The conclusion and future work are finally given in Section~\ref{sec:Conclusion and future work}.

\section{Description of Chinese character caption}
\label{sec:Description of Chinese character caption}
\begin{figure}
\centering
\includegraphics[width=2.6in]{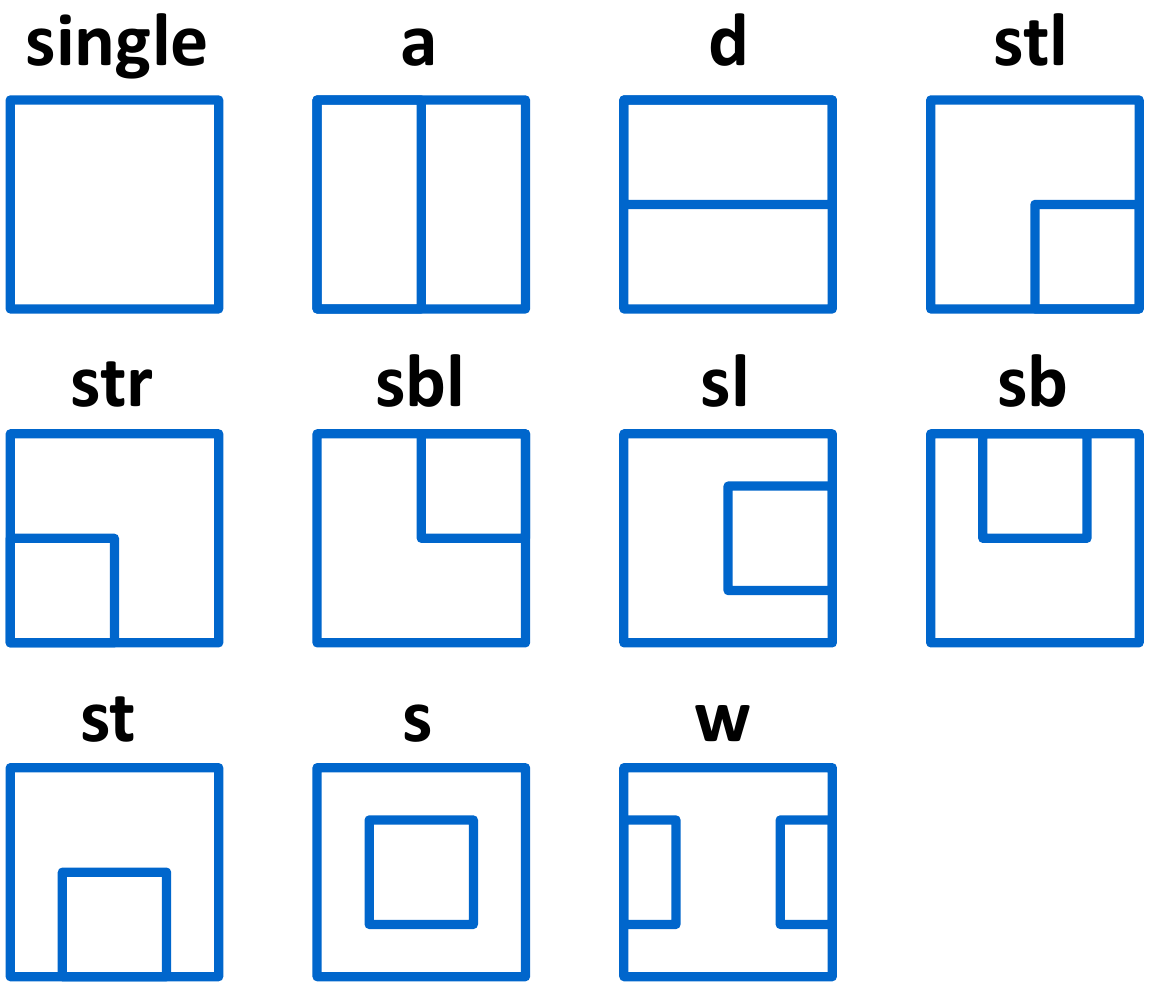}
\caption{Graphical representation of eleven common spatial structures among radicals, different radicals are divided by internal line.}
\label{fig:radical-structure}
\end{figure}

In this section, we will introduce how we generate captions of Chinese characters. The character caption is composed of three key components: radicals, spatial structures and a pair of braces (e.g. ``\{'' and ``\}''). A radical represents a basic part of Chinese character and it is often shared by different Chinese characters. Compared with enormous Chinese character categories, the amount of radicals is quite limited. It is declared in GB13000.1 standard published by National Language Committee of China that nearly 500 radicals consist of over 20,000 Chinese characters. As for the complicated two-dimensional spatial structures among radicals, Fig.~\ref{fig:radical-structure} illustrates eleven common structures and the descriptions are demonstrated as follows:
\begin{itemize}
  \item \textbf{single-element:} sometimes a single radical represents a Chinese character and therefore we can not find internal structures in such characters
  \item \textbf{a:} left-right structure
  \item \textbf{d:} top-bottom structure
  \item \textbf{stl:} top-left-surround structure
  \item \textbf{str:} top-right-surround structure
  \item \textbf{sbl:} bottom-left-surround structure
  \item \textbf{sl:} left-surround structure
  \item \textbf{sb:} bottom-surround structure
  \item \textbf{st:} top-surround structure
  \item \textbf{s:} surround structure
  \item \textbf{w:} within structure
\end{itemize}

After decomposing Chinese characters into radicals and internal spatial structures by following~\cite{cjk-decomp}, we use a pair of braces to constrain a single structure. Take ``stl'' as an example, it is captioned as ``stl \{ radical-1 radical-2 \}''. The generation of a Chinese character caption is finished when all radicals are included in the caption.

\section{The proposed approach}
\label{sec:The proposed approach}

\begin{figure}
\centering
\includegraphics[width=3.2in]{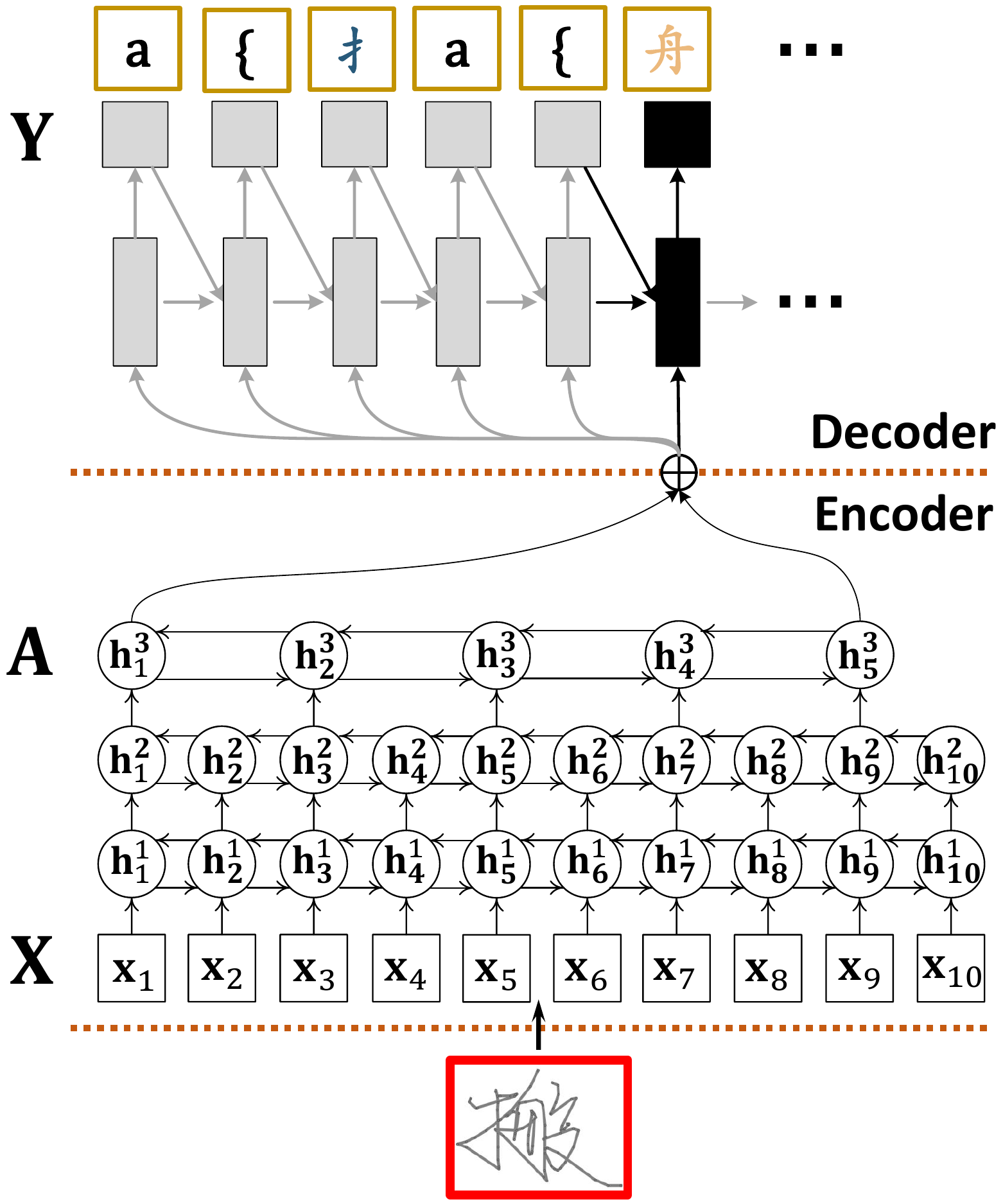}
\caption{Overall framework of TRAN for online handwritten Chinese character recognition. It is composed of a bidirectional RNN encoder and a unidirectional RNN decoder.}
\label{fig:TRAN-framework}
\end{figure}

In this section, we elaborate the proposed TRAN framework, namely generating an underlying Chinese character caption from a sequence of online handwritten trajectory points, as illustrated in Fig.~\ref{fig:TRAN-framework}. First, we extract trajectory information as the input feature from original trajectory points (xy-coordinates). A stack of bidirectional RNNs are then employed as the encoder to transform the input feature into high-level representations. Since the original trajectory points are a variable-length sequence, the extracted high-level representations are also a variable-length sequence. To associate the variable-length representations with variable-length character caption, we generate a fixed-length context vector via weighted summing the high-level representations and a unidirectional RNN decoder uses the fixed-length context vector to generate the character caption one symbol at a time. We introduce an attention model to produce the weighting coefficients so that the context vector can contain only useful trajectory information at each decoding step.

\subsection{Feature extraction}
\label{sec:Feature extraction}
During the data acquisition of online handwritten Chinese character, the pen-tip movements (xy-coordinates) and pen states (pen-down or pen-up) are stored as variable-length sequential data:
\begin{equation}\label{eq:trajectory input}
\left \{\left[ {{x_1},{y_1},{s_1}} \right],\;\left[ {{x_2},{y_2},{s_2}} \right],\; \ldots \;,\;\left[ {{x_N},{y_N},{s_N}} \right] \right\}
\end{equation}
where $N$ is the length of sequence, ${x_i}$ and ${y_i}$ are the xy-coordinates of the pen movements and ${s_i}$ indicates which stroke the $i^{\textrm{th}}$ point belongs to.

To address the issue of non-uniform sampling by different writing speed and the size variations of the coordinates on different potable devices, the interpolation and normalization to the original trajectory points are first conducted according to~\cite{zhang2017drawing}. Then we extract a 6-dimensional feature vector for each point:
\begin{equation}\label{eq:6dimensional feature}
\left[ {{x_i},{y_i},\Delta {x_i},\Delta {y_i},\delta ({s_i} = {s_{i + 1}}),\delta ({s_i} \ne {s_{i + 1}})} \right]
\end{equation}
where $\Delta {x_i}={x_{i+1}}-{x_i}$, $\Delta {y_i}={y_{i+1}}-{y_i}$, and $\delta ( \cdot ) = 1$ when the condition is true or zero otherwise. The last two terms are flags which indicate the status of the pen, i.e., $\left[ {1,\;0} \right]$ and $\left[ {0,\;1} \right]$ are pen-down and pen-up respectively. For convenience, in the following sections, we use $\mathbf{X}=\left( {{\mathbf{x}_1},\;{\mathbf{x}_2},\; \ldots ,\;{\mathbf{x}_N}} \right)$ to denote the input sequence of the encoder, where ${\mathbf{x}_i} \in {\mathbb{R}^d}$ ($d=6$).

\subsection{Encoder}
\label{sec:Encoder}
Given the feature sequence $\left( {{\mathbf{x}_1},\;{\mathbf{x}_2},\; \ldots ,\;{\mathbf{x}_N}} \right)$, we employ RNN as the encoder to encode them into high-level representations as RNN has shown its strength in processing sequential signals. However, a simple RNN has revealed serious problems during training namely vanishing gradient and exploding gradient~\cite{bengio1994learning,zhang2016rnn}. Therefore, an improved version of RNN named gated recurrent units (GRU)~\cite{chung2014empirical} which can alleviate these two problems is employed in this study as it utilizes an update gate and a reset gate to control the flow of forward information and backward gradient. The GRU hidden state ${\mathbf{h}_t}$ in encoder is computed by:
\begin{equation}\label{eq:GRU function}
{{\mathbf{h}}_t} = \textrm{GRU} \left( {{\mathbf{x}_t}, {\mathbf{h}_{t - 1}}} \right)
\end{equation}
and the GRU function can be expanded as follows:
\begin{align}\label{eq:expand GRU}
 & {{\mathbf{z}}_t} = \sigma ({{\mathbf{W}}_{xz}}{{\mathbf{x}}_{t}} + {{\mathbf{U}}_{hz}}{{\mathbf{h}}_{t - 1}}) \\
 & {{\mathbf{r}}_t} = \sigma ({{\mathbf{W}}_{xr}}{{\mathbf{x}}_{t}} + {{\mathbf{U}}_{hr}}{{\mathbf{h}}_{t - 1}}) \\
 & {{\bf{\tilde h}}_t} = \tanh ({{\bf{W}}_{xh}}{{\bf{x}}_{t}} + {{\bf{U}}_{rh}}({{\bf{r}}_t} \otimes {{\bf{h}}_{t - 1}})) \\
 & {{\bf{h}}_t} = (1 - {{\bf{z}}_t}) \otimes {{\bf{h}}_{t - 1}} + {{\bf{z}}_t} \otimes {{\bf{\tilde h}}_t}
\end{align}
where $\sigma$ denotes the sigmoid activation function, $\otimes$ denotes an element-wise multiplication operator, ${{\mathbf{z}}_t}$, ${{\mathbf{r}}_t}$ and ${{\bf{\tilde h}}_t}$ are the update gate, reset gate and candidate activation, respectively. ${\mathbf{W}}_{xz}$, ${\mathbf{W}}_{xr}$, ${\bf{W}}_{xh}$, ${\mathbf{U}}_{hz}$, ${\mathbf{U}}_{hr}$ and ${\bf{U}}_{rh}$ are related weight matrices.

Nevertheless, even if the unidirectional GRU can have access to the history of input signals, it does not have the ability of modeling future context. Therefore we exploit the bidirectional GRU by passing the input vectors through two GRU layers running in opposite directions and concatenating their hidden state vectors so that the encoder can use both history and future information. To obtain a high-level representation, the encoder stacks multiple GRU layers on top of each other as illustrated in Fig.~\ref{fig:TRAN-framework}. In this study, our encoder consists of 4 bidirectional GRU layers. Each layer has 250 forward and 250 backward GRU units. We also add pooling over time axes in high-level GRU layers because: 1) the high-level representations are overly precise and contain much redundant information; 2) the decoder needs to attend less if the number of encoder output reduces, leading to improvement of performance; 3) the pooling operation accelerates the encoding process. The pooling is applied to the top GRU layer by dropping the even output over time.

Assuming the bidirectional GRU encoder produces a high-level representation sequence ${\mathbf{A}}$ with length ${L}$. Because there is one pooling operation in the bidirectional GRU encoder, $L = \frac{N}{2}$. Each of these representations is a $D$-dimensional vector ($D=500$):
\begin{equation}\label{eq:annotation A}
  \mathbf{A} = \left\{ {{{\mathbf{a}}_1}, \ldots ,{{\mathbf{a}}_L}} \right\}\;,\;{{\mathbf{a}}_i} \in {\mathbb{R}^D}
\end{equation}

\subsection{Decoder with attention model}
\label{sec:Decoder with attention model}
After obtaining high-level representations ${\mathbf{A}}$, the decoder aims to make use of them to generate a Chinese character caption. The output sequence ${\mathbf{Y}}$ is represented as a sequence of one-hot encoded vectors:
\begin{equation}\label{eq:caption Y}
 \mathbf{Y} = \left\{ { \mathbf{y}_1, \ldots ,\mathbf{y}_C} \right\}\;,\;{{\mathbf{y}}_i} \in {\mathbb{R}^K}
\end{equation}
where $K$ is the vocabulary size and $C$ is the length of character caption. Note that, both the length of representation sequence (${L}$) and the length of character caption (${C}$) are variable. To address the mapping from variable-length representation sequence to variable-length character caption, we attempt to compute an intermediate fixed-size vector ${\mathbf{c}_t}$ that incorporates useful information of representation sequence. The decoder then utilizes this fixed-size vector to predict the character caption one symbol at a time. As ${\mathbf{c}_t}$ contains overall information of input sequence, we call it context vector. At each decoding step, the probability of the predicted word is computed by the context vector ${\mathbf{c}_t}$, current decoder state ${\mathbf{s}_t}$ and previous predicted symbol ${\mathbf{y}_{t-1}}$ using a multi-layer perceptron:
\begin{equation}\label{eq:compute Y}
  p({{\mathbf{y}}_t}|{{\mathbf{y}}_{t - 1}},{\mathbf{X}}) = g \left ({{\mathbf{W}}_o}h({\mathbf{E}}{{\mathbf{y}}_{t - 1}} + {{\mathbf{W}}_s}{{\mathbf{s}}_t} + {{\mathbf{W}}_c}{{\mathbf{c}}_t})\right )
\end{equation}
where $g$ denotes a softmax activation function over all the symbols in the vocabulary, $h$ denotes a maxout activation function. Let $m$ and $n$ denote the dimensions of embedding and decoder state, ${{\mathbf{W}}_o} \in {\mathbb{R}^{K \times \frac{m}{2}}}$, ${{\mathbf{W}}_s} \in {\mathbb{R}^{m \times n}}$, ${{\mathbf{W}}_c} \in {\mathbb{R}^{m \times D}}$, and ${\mathbf{E}}$ denotes the embedding matrix.

Since the context vector ${\mathbf{c}_t}$ needs to be fixed-length, it is an intuitive way to produce it by summing all representation vectors ${\mathbf{a}_i}$ at time step $t$. However, average summing is too robust and leads to loss of useful information. Therefore, we adopt weighted summing while the weighting coefficients are called attention probabilities. The attention probability performs as a description that tells which part of representation sequence is useful at each decoding step. We compute the decoder state ${\mathbf{s}_t}$ and context vector ${\mathbf{c}_t}$ as follows:
\begin{align}\label{eq:decoder with attention}
 & {{\mathbf{\hat s}}_t} = \textrm{GRU} \left( {{\bf{y}}_{t-1}}, {{\bf{s}}_{t - 1}} \right) \\
 & {\mathbf{F}} = {\mathbf{Q}} * \sum\nolimits_{l=1}^{t - 1} {{{\bm{\alpha}}_l}} \label{eq:coverage} \\
 & {e_{ti}} = {\bm{\nu }}_{\text{att}}^{\rm T}\tanh ({{\mathbf{W}}_{\text{att}}}{{\mathbf{\hat s}}_t} + {{\mathbf{U}}_{\text{att}}}{{\mathbf{a}}_i} + {{\mathbf{U}}_f}{{\mathbf{f}}_i}) \\
 & {\alpha _{ti}} = \frac{{\exp ({e_{ti}})}}{{\sum\nolimits_{k = 1}^L {\exp ({e_{tk}})} }} \\
 & {{\mathbf{c}}_t} = \sum\nolimits_{i=1}^L {{\alpha _{ti}}{{\mathbf{a}}_i}} \\
 & {{\mathbf{s}}_t} = \textrm{GRU} \left( {{\mathbf{c}}_t}, {{\mathbf{\hat s}}_t} \right)
\end{align}
Here, we can see that the decoder adopts two unidirectional GRU layers to calculate the decoder state ${\mathbf{s}_t}$. The GRU function is the same one in Eq.~\eqref{eq:GRU function}. ${{\mathbf{\hat s}}_t}$ denotes the current decoder state prediction, ${{e}_{ti}}$ denotes the energy of ${\mathbf{a}_i}$ at time step $t$ conditioned on ${\mathbf{\hat s}_t}$. The attention probability ${{\alpha}_{ti}}$, which is the $i^{\text{th}}$ element of $\bm{\alpha}_t$, is computed by taking ${{e}_{ti}}$ as input of a softmax function. The context vector ${\mathbf{c}_t}$ is then calculated via weighted summing representation vectors ${\mathbf{a}_i}$ with attention probabilities employed as weighting coefficients. During the computation of attention probability, we also append a coverage vector ${\mathbf{f}_i}$ (the $i^{\text{th}}$ vector of $\mathbf{F}$) in the attention model. The coverage vector is computed based on the summation of all past attention probabilities so that the coverage vector contains the information of alignment history as shown in Eq.~\eqref{eq:coverage}. We adopt the coverage vector in order to let the attention model know which part of representation sequence has been attended or not~\cite{tu2016modeling}. Let $n^{'}$ denote the attention dimension. Then ${{\bm{\nu }}_{\text{att}}} \in {\mathbb{R}^{{n^{'}}}}$, ${{\mathbf{W}}_{\text{att}}} \in {\mathbb{R}^{{n^{'}} \times n}}$ and ${{\mathbf{U}}_{\text{att}}} \in {\mathbb{R}^{{n^{'}} \times D}}$.

%The coverage vector is expected to adjust the future attention. More specifically, trajectory points in the input sequence already significantly contributed to the generation of target symbols should be assigned with lower attention probabilities in the following decoding phases. On the contrary, trajectory points with less contributions should be assigned with higher attention probabilities. Consequently, the decoding process is finished only when the entire input sequence has contributed and the problems of over-translating or under-translating can be alleviated.

\section{Training and Testing Details}
\label{sec:Training and Testing Details}

The training objective of the proposed model is to maximize the predicted symbol probability as shown in Eq.~\eqref{eq:compute Y} and we use cross-entropy (CE) as the objective function:
\begin{equation}\label{eq:objective}
  O = - \sum\nolimits_{t=1}^C \log p({w_t}|{\mathbf{y}_{t-1},\mathbf{X}})
\end{equation}
where $w_t$ represents the ground truth word at time step $t$, $C$ is the length of output string. The implementation details of GRU encoder has been introduced in Section~\ref{sec:Encoder}. The decoder uses two layers with each using 256 forward GRU units. The embedding dimension $m$, decoder state dimension $n$ and attention dimension $n'$ are all set to 256. The convolution kernel size for computing coverage vector is set to $(5 \times 1)$ as it is a one-dimensional convolution operation, while the number of convolution filters is set to 256. We utilize the adadelta algorithm~\cite{zeiler2012adadelta} with gradient clipping for optimization. The adadelta hyperparameters are set as $\rho = 0.95$, $\varepsilon = {10^{ - 8}}$.

In the decoding stage, we aim to generate a most likely character caption given the input trajectory:
\begin{equation}\label{eq:decoding objective}
  {\mathbf{\hat y}} = \mathop {\arg \max }\limits_{\mathbf{y}} \log P\left( {{\mathbf{y}}|{\mathbf{X}}} \right)
\end{equation}
However, different from the training procedure, we do not have the ground truth of previous predicted word. To prevent previous prediction errors inherited by next decoding step , a simple left-to-right beam search algorithm~\cite{cho2015natural} is employed to implement the decoding procedure. Here, we maintained a set of 10 partial hypotheses beginning with the start-of-sentence $<sos>$. At each time step, each partial hypothesis in the beam is expanded with every possible word and only the 10 most likely beams are kept. This procedure is repeated until the output word becomes the end-of-sentence $<eos>$.

\section{Experiments}
\label{sec:Experiments}
In this section, we present experiments on recognizing seen and unseen online handwritten Chinese character classes by answering the following questions:
\begin{description}
    \item[Q1] Is the TRAN effective when recognizing seen Chinese character classes?
    \item[Q2] Is the TRAN effective when recognizing unseen Chinese character classes?
    \item[Q3] How does the TRAN analyze the radicals and spatial structures?
\end{description}

\subsection{Performance on recognition of seen Chinese character classes (Q1)}
\label{sec:Performance on recognition of seen Chinese character classes (Q1)}
In this section, we show the effectiveness of TRAN on recognizing seen Chinese character classes. The set of character class is 3,755 commonly used Chinese characters. The dataset used for training is the CASIA~\cite{liu2011casia} dataset including OLHWDB1.0 and OLHWDB1.1. There are totally 2,693,183 samples for training and 224,590 samples for testing. The training and testing data were produced by different writers with enormous handwriting styles across individuals.
\begin{table}[h]
\caption{\label{tab:1}{Results on CASIA dataset of online handwritten Chinese character recognition.}}
\centering
\begin{tabular}{l c c}
\toprule
\textbf{Methods} & \textbf{Reference} & \textbf{Accuracy}\\
\midrule
Human Performance & \cite{yin2013icdar} & 95.19\%\\
Traditional Benchmark & \cite{liu2013online} & 95.31\%\\
NET4 & \cite{zhang2017drawing} & 96.03\%\\
TRAN & -- & 96.43\%\\
\bottomrule
\end{tabular}
\end{table}
In Table~\ref{tab:1}, the human performance on CASIA test set and the previous benchmark are both listed. NET4 is the proposed method in~\cite{zhang2017drawing} which represents the state-of-the-art method on CASIA dataset and it belongs to non-radical based methods. NET4 achieved an accuracy of 96.03\% while TRAN achieved an accuracy of 96.43\%, revealing relative character error rate reduction of 10\%. To be fairly comparable, here NET4 and TRAN both did not use the sequential dropout trick as proposed in~\cite{zhang2017drawing} so the performance of NET4 is not as good as the best performance presented in~\cite{zhang2017drawing}. As explained in the contributions of this study in Section~\ref{sec:Introduction}, the main difference between radical based method and non-radical based method for Chinese character recognition is the size of radical vocabulary is largely less than Chinese character vocabulary, yielding decrease of redundancy among output classes and improvement of recognition performance.

\subsection{Performance on recognition of unseen Chinese character classes (Q2)}
\label{sec:Performance on recognition of unseen Chinese character classes (Q2)}
The number of Chinese character classes is not fixed as more and more novel characters are being created. Also, the overall Chinese character classes are enormous and it is difficult to train a recognition system that covers them all. Therefore it is necessary for a recognition system to possess the capability of recognizing unseen Chinese characters, called zero-shot learning.

Obviously traditional non-radical based methods are incapable of recognizing these unseen characters since the objective character class has never been seen during training procedure. However TRAN is able to recognize unseen Chinese characters only if the radicals composing unseen characters have been seen. To validate the performance of TRAN on recognizing unseen Chinese character classes, we divide 3755 common Chinese characters into 3255 classes and the other 500 classes. We choose handwritten characters belonging to 3255 classes from original training set as the new training set and we choose handwritten characters belonging to the other 500 classes from original testing set as the new testing set. By doing so, both the testing character classes and handwriting variations have never been seen during training. We explore different size of training set to train TRAN, ranging from 500 to 3255 Chinese character classes and we make sure the radicals of testing characters are covered in training set.
\begin{table}[h]
\caption{\label{tab:2}{Results on newly divided testing set based on CASIA dataset of online handwritten unseen Chinese character recognition.}}
\centering
\begin{tabular}{c c c}
\toprule
\textbf{Train classes} & \textbf{Train samples} & \textbf{Test Accuracy}\\
\midrule
500 & 359,036 & -\\
1000 & 717,194 & 10.74\%\\
1500 & 1,075,344 & 26.02\%\\
2000 & 1,435,295 & 39.35\%\\
2755 & 1,975,972 & 50.45\%\\
3255 & 2,335,433 & 60.37\%\\
\bottomrule
\end{tabular}
\end{table}

We can see in Table~\ref{tab:2} the recognition accuracy of unseen Chinese character classes is not available when training set only contains 500 Chinese character classes. We believe it is difficult to train TRAN properly to accommodate large handwriting variations when the number of character classes is quite small. When the training set contains 3255 character classes, TRAN achieves a character accuracy of \textbf{60.37}\% which is a relatively pleasant performance compared with traditional recognition systems as they can not recognize unseen Chinese character classes which means their accuracies are definitely \textbf{0}\%. The performance of recognizing unseen Chinese character classes is not as good as the performance presented in~\cite{zhang2017ran} because the handwritten Chinese characters are much more ambiguous compared with printed Chinese characters due to the large handwriting variations.

\subsection{Attention visualization (Q3)}
\label{sec:Attention visualization (Q3)}

\begin{figure}
\centering
\includegraphics[width=2.8in]{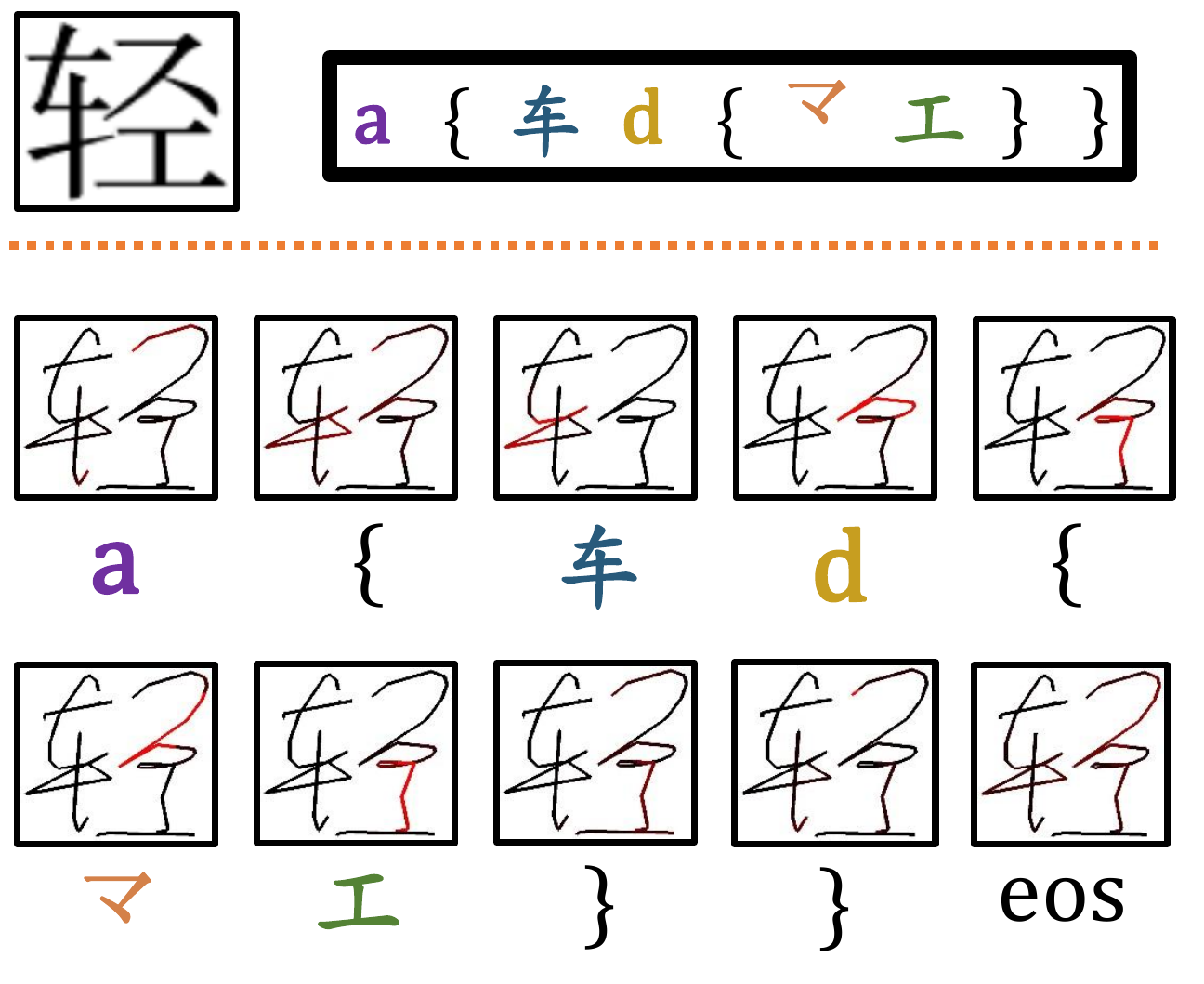}
\caption{Examples of attention visualization during the decoding procedure. The red color on trajectory describes the attention probabilities namely the lighter color denotes higher attention probabilities and the darker color denotes lower attention probabilities.}
\label{fig:attention_visualization}
\end{figure}

In this section, we show through attention visualization how TRAN is able to recognize internal radicals and analyze the two-dimensional spatial structure among radicals. Fig.~\ref{fig:attention_visualization} illustrates an example of attention visualization. Above the dotted line, there is one Chinese character class and its corresponding character caption. Below dotted line, there are images denoting the visualization of attention probabilities during decoding procedure. We draw the trajectory of input handwritten Chinese character in a two-dimensional greyscale image to visualize attention. Below images there are corresponding symbols generated by decoder at each decoding step.

As we can see in Fig.~\ref{fig:attention_visualization}, when encountering basic radicals, the attention model generates the alignment well corresponding to the human intuition. Also, it mainly focus on the ending of last radical and the beginning of next radical to detect a spatial structure. Take ``d'' as an example, by attending to the ending of last radical and the beginning of next radical, the attention model detects a top-bottom direction, therefore a top-bottom structure is analyzed. Immediately after generating a spatial structure, the decoder produces a pair of braces ``\{\}'', which are employed to constrain the two-dimensional structure in Chinese character caption.

\section{Conclusion and future work}
\label{sec:Conclusion and future work}
In this study we introduce TRAN for online handwritten Chinese character recognition. The proposed TRAN recognizes Chinese character by identifying internal radicals and analyzing spatial structures among radicals. We show from experimental results that TRAN outperforms the state-of-the-art method on recognition of online handwritten Chinese characters and possesses the capability of recognizing unseen Chinese character categories. By visualizing learned attention probabilities, we can observe the alignments of radicals and analysis of structures correspond well to human intuition.

%In future work, we will explore the postprocessing of TRAN for recognizing seen Chinese characters as the exact matching of ground-truth character caption is too strict. We also plan to improve the capability of TRAN for recognizing unseen Chinese character classes. The rule of generating Chinese character captions also need to be modified.

%\section*{Acknowledgment}
%This work was supported in part by the National Natural Science Foundation of China under Grants 61671422 and U1613211, in part by the National Key Research and Development Program of China under Grant 2017YFB1002200, in part by the MOE-Microsoft Key Laboratory of USTC.

\bibliographystyle{IEEEtran}
\bibliography{refs}

\end{document}